\documentclass[lettersize,journal]{IEEEtran}
\usepackage{amsmath,amsfonts}
\usepackage{algorithmic}
\usepackage{algorithm}
\usepackage{array}
\usepackage[caption=false,font=normalsize,labelfont=sf,textfont=sf]{subfig}
\usepackage{textcomp}
\usepackage{stfloats}
\usepackage{url}
\usepackage{verbatim}
\usepackage{graphicx}
\usepackage{cite}
\usepackage{comment}
\begin{document}

\title{Local-Global Context-Aware and Structure-Preserving Image Super-Resolution}





\author{
    Sanchar Palit, Subhasis Chaudhuri, and Biplab Banerjee\\
    Indian Institute of Technology Bombay, India\\
}






\maketitle

\begin{abstract}
Diffusion models have recently achieved significant success in various image manipulation tasks, including image super-resolution and perceptual quality enhancement. Pretrained text-to-image models, such as Stable Diffusion, have exhibited strong capabilities in synthesizing realistic image content, which makes them particularly attractive for addressing super-resolution tasks. While some existing approaches leverage these models to achieve state-of-the-art results, they often struggle when applied to diverse and highly degraded images, leading to noise amplification or incorrect content generation. To address these limitations, we propose a contextually precise image super-resolution framework that effectively maintains both local and global pixel relationships through Local-Global Context-Aware Attention, enabling the generation of high-quality images. Furthermore, we propose a distribution- and perceptual-aligned conditioning mechanism in the pixel space to enhance perceptual fidelity. This mechanism captures fine-grained pixel-level representations while progressively preserving and refining structural information, transitioning from local content details to the global structural composition. During inference, our method generates high-quality images that are structurally consistent with the original content, mitigating artifacts and ensuring realistic detail restoration. Extensive experiments on multiple super-resolution benchmarks demonstrate the effectiveness of our approach in producing high-fidelity, perceptually accurate reconstructions.
\end{abstract}


\begin{figure*}
  \centering
   \includegraphics[width=1.0\linewidth]{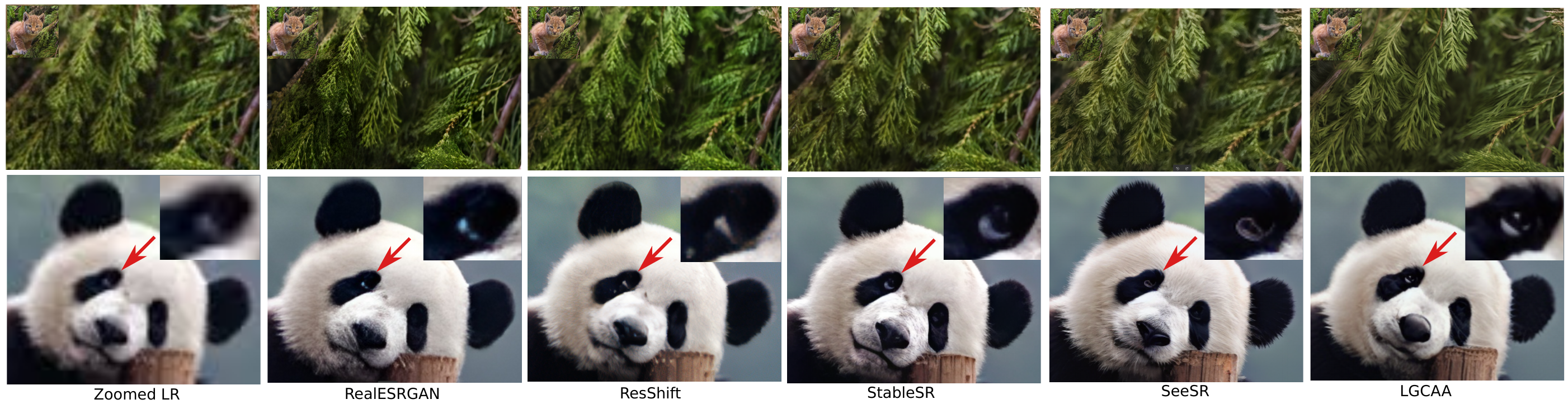}

   \caption{Comparison of our method with recent state-of-the-art approaches on a degraded image. While existing methods introduce high-frequency details, they often deviate from the original content. In contrast, our method produces high-quality images that maintain a realistic appearance at a global scale while preserving fidelity to the original content when examined closely. This balance between high-frequency and low-frequency information ensures a more natural reconstruction. Please zoom in for a detailed view.
   }
   \label{fig:teaser_comparison}
\end{figure*}

\section{Introduction}
\label{sec:intro}
Image super-resolution is a challenging task due to the degradation process, which leads to the loss of essential image information, making accurate reconstruction difficult. This degradation can be modeled as individual effects such as blurring and noise addition or as a combination of multiple factors. Early research in this field assumed predefined image degradations and developed various methods~\cite{chen2021pre,dong2014learning,ma2106text,sun2024perception,zhang2022efficient,zhang2018image} to address the problem. However, these approaches are limited in their ability to achieve high-fidelity image reconstruction and struggle to handle extreme degradation scenarios effectively.

With the advent of generative models such as Generative Adversarial Networks (GAN)~\cite{goodfellow2020generative} have been employed to model the degradation process~\cite{bulat2018learn} through adversarial training, enabling the reconstruction of high-quality images by approximating the reverse transformation. GAN-based methods~\cite{chen2023human,liang2022details,liang2022efficient,xie2023desra} have been particularly effective in generating perceptually high-quality images under complex degradation conditions. Additionally, datasets containing large-scale low-resolution (LR) and high-resolution (HR) image pairs~\cite{cai2019toward,wei2020component,wang2021real} have been introduced, encompassing various real-world degradations to facilitate more effective and standardized evaluation which formulates the problem of Real world Image Super-Resolution (Real-ISR) to remove possible real world complex degradation.

\begin{figure}
    \centering
    \includegraphics[width=1.0\linewidth]{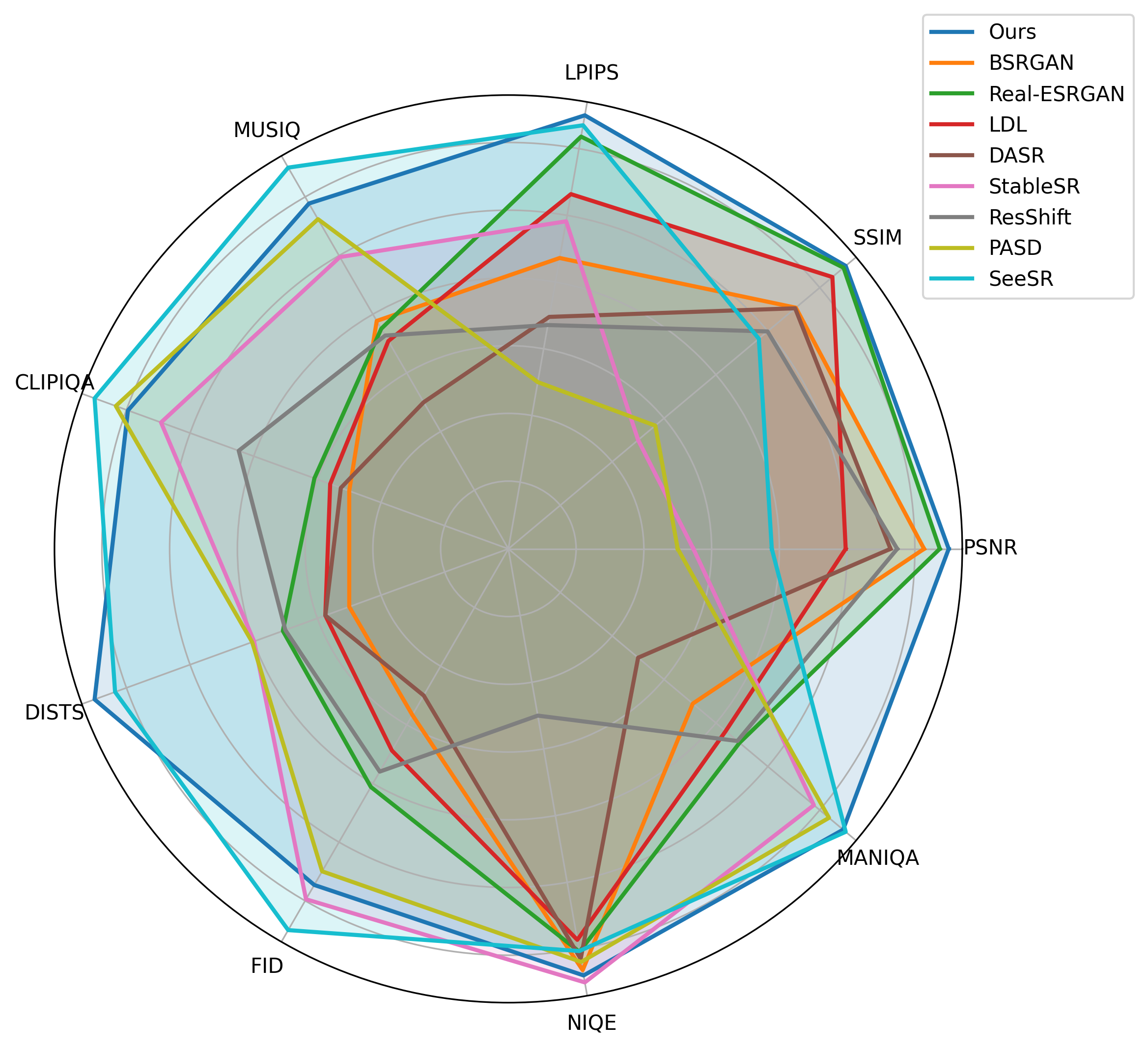}
    \caption{Comparison of performance and efficiency among Real-ISR methods. For visualization, the metrics LPIPS, DISTS, FID, and NIQE, which are lower-is-better measures of image quality, are inverted and normalized. The proposed method attains superior performance across the majority of evaluated metrics.}
    \label{fig:metric_radar}
\end{figure}

Approaches such as BSRGAN~\cite{zhang2021designing} and Real-ESRGAN~\cite{wang2021real} have demonstrated significant improvements, producing results with enhanced detail and realism. However, GAN-based models still have several limitations, including the introduction of noise, suppression of original content with artificially generated details, and in some cases, the amplification of undesired artifacts from the LR input, leading to inaccurate reconstructions.

The introduction of diffusion models~\cite{sohl2015deep,ho2020denoising} for image generation has alleviated the challenges associated with the complex training process of GANs.  The diffusion process can follow a Markov chain-based Denoising Diffusion Probabilistic Model (DDPM)~\cite{ho2020denoising,kingma2021variational} or utilize Stochastic Differential Equations (SDEs) in combination with score matching networks~\cite{song2020score,song2020improved,song2021maximum} to estimate and remove noise. Additionally, diffusion models have facilitated Real-ISR~\cite{saharia2022image} and other image restoration tasks by enabling conditioning on various modalities, such as text, LR images, or image-specific features~\cite{ramesh2021zero,rombach2022high,zhang2023adding} like edge maps and high-frequency details. ResShift~\cite{yue2023resshift} has emerged as a notable approach, leveraging stepwise error shifting within the diffusion framework to progressively refine LR images into HR counterparts. Furthermore, the introduction of ControlNet~\cite{zhang2023adding} has allowed for spatially conditioned diffusion processes by incorporating different image-based features, such as edges, and other high-level attributes. The advancement of text-to-image models~\cite{ramesh2021zero,ramesh2022hierarchical,ding2021cogview,nichol2021glide,saharia2022photorealistic}, particularly diffusion-based approaches such as Stable Diffusion~\cite{rombach2022high}, has paved new pathways for Real-ISR. Trained on large-scale datasets, these models have learned realistic image formation principles from textual descriptions, enabling applications in image editing, inpainting, and various forms of conditional image manipulation—either from pure noise or an initial degraded image. Building on these advancements, works such as StableSR~\cite{wang2024exploiting}, SeeSR~\cite{wu2024seesr}, and DiffBIR~\cite{lin2024diffbir} have emerged for real-world ISR tasks. StableSR and DiffBIR utilize diffusion priors to enhance super-resolution performance, while SeeSR is specifically trained to extract semantic prompts from LR images. By leveraging the semantic understanding inherent in diffusion models, SeeSR aims to maintain text-based relationships in the super-resolution process. However, as the method relies on text-based semantic conditioning, it is prone to generating unintended artifacts when the degradation in the input image is severe.

We propose a model for Real-ISR that harnesses the well-trained image formation capabilities of Stable Diffusion while ensuring the effective preservation of the contextual information present in the LR image. Any global structure or texture can be represented locally, and our method ensures that local edges are preserved, sharpened, and refined, thereby maintaining and enhancing the overall global texture and structure. To achieve this, we integrate the LR conditioning image into the Stable Diffusion pipeline using a Local-Global Context Aware Attention (LGCAA) module. This module ensures the preservation of local region relationships while enabling individual pixels to capture long-range dependencies through global attention mechanisms. In addition, we introduce the Distribution and Perceptual Aligned Conditioning Module (DPACM) to preserve the structural consistency between LR and HR images while ensuring effective histogram preservation in the latent space. This module is designed to maintain the perceptual quality of the generated HR images. To achieve this, we employ the Wasserstein distance to align the pixel distributions of the LR and HR images, ensuring faithful reconstruction. Furthermore, we incorporate a perceptual loss, leveraging a robust ControlNet-based feature extractor to enhance the perceptual quality of the output. During inference, our model is capable of generating high-quality and high-fidelity images by preserving the content of the LR input while significantly improving visual quality as shown in Fig.~\ref{fig:teaser_comparison}. Experimental results demonstrate that the proposed Real-ISR model achieves consistently strong performance across diverse scene contents, generating perceptually appealing super-resolved images, as illustrated in Figure~\ref{fig:metric_radar}.




\section{Related Work}
\subsection{GAN based Real-ISR}
Adversarial training-based methods, which enable image generation from pure noise, have been successfully applied to Real-ISR~\cite{zhang2021designing,liang2022details,liang2022efficient,chen2022real} to handle complex degradations, surpassing conventional deep learning techniques~\cite{chen2023dual,chen2023activating,dai2019second,liang2021swinir,lim2017enhanced, liu2019hierarchical,haris2018deep}. Pioneering works such as BSRGAN~\cite{zhang2021designing} and Real-ESRGAN~\cite{wang2021real} have demonstrated that image restoration becomes significantly more effective even in severe degradation scenarios through adversarial training. Subsequently, methods like LDL~\cite{liang2022details} and DASR~\cite{liang2022efficient} have further improved results by focusing on artifact detection and removal while enhancing image details for better restoration. However, despite their effectiveness, GAN-based techniques suffer from challenging and computationally intensive training processes. Moreover, conditioning the image generation process on multiple modalities remains a challenging task. Additionally, the model is still prone to mode collapse, which can result in suboptimal restoration in certain scenarios.

\subsection{Diffusion models and Diffusion prior based Real-ISR}
Diffusion models have demonstrated remarkable efficiency in image synthesis~\cite{ho2020denoising,rombach2022high} and are primarily formulated using a Markov chain framework. Another variant of diffusion models, based on stochastic differential equations (SDEs) with score-based networks~\cite{song2020improved,song2020score,song2021maximum}, has also been utilized for training text-conditioned diffusion models. Initially, diffusion models operated in pixel space, but with the introduction of latent diffusion models~\cite{rombach2022high}, they have been adapted to the latent space, enabling HR image generation while processing in a lower-dimensional space. DDPMs have been employed in ResShift~\cite{yue2023resshift}, utilizing different noise scheduling strategies to progressively refine low-quality images into high-quality ones by iteratively shifting noise residuals. DDPM-based text-to-image models~\cite{rombach2022high,ramesh2021zero,ding2021cogview,nichol2021glide,saharia2022photorealistic} have been trained on extensive natural image datasets with text conditioning to generate high-fidelity images. It has been observed that the conditioning mechanism can extend beyond text, allowing guidance through various modalities to direct the diffusion process toward specific outputs. Efficient sampling strategies, including DDIM and other distillation-based techniques, have been introduced to generate high-quality images within a reduced number of diffusion steps. Building on these advancements, we propose an Real-ISR method leveraging the text-to-image Stable Diffusion model~\cite{rombach2022high} to guide the transformation from LR to HR images using purely image-based features, effectively detaching the process from text embeddings. To capture both local pixel relationships and long-range dependencies, we integrate a Local and Global Context-Aware Attention mechanism to enhance the Stable Diffusion model, ensuring high-fidelity image reconstruction with improved structural consistency.

\begin{figure*}[!ht]
  \centering
   \includegraphics[width=1.0\linewidth]{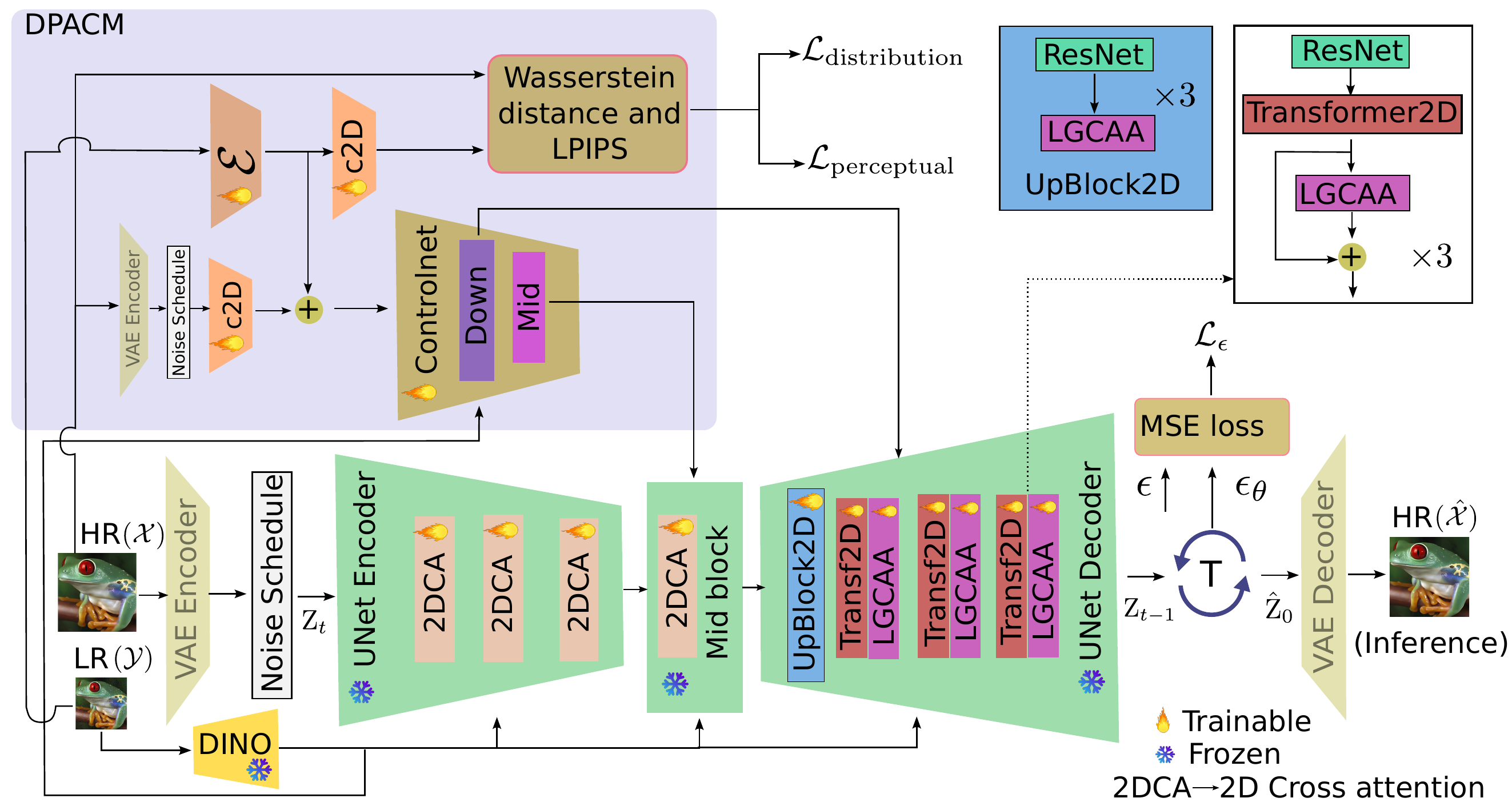}

   \caption{Overall architecture of the proposed LGCAA during training. Here $\mathcal{E}$ is the controlnet conditional embedding and c2D is the 2Dconvolutional block. During training, the model is optimized in the latent space, where the U-Net remains partially frozen with only its attention blocks being trainable. In parallel, the 2D U-Net is conditioned through a trainable ControlNet~\cite{zhang2023adding}, integrated via zero-convolution layers and further augmented with image features extracted from an additional frozen DINO module. During inference, the conditioning part of the Unet and the controlnet is not used instead, the LR is used as input to get the HR image.}
   \label{fig:main_diag}
\end{figure*}

\section{Method}
\label{sec:method}
\subsection{Problem formulation}
During the degradation process, an image $\mathcal{X}$ undergoes a degradation operation $\mathcal{D}$, resulting in a LR image $\mathcal{Y} =\mathcal{D}(\mathcal{X})$. This degradation process may consist of a single transformation or a combination of multiple degradations, such as $ \mathcal{D} = \{\mathcal{D}_1, \mathcal{D}_2, ..\}$. In diffusion model-based image restoration, the degradation process is typically modeled as a combination of Gaussian noise perturbations. The restoration process then involves estimating and subsequently removing the Gaussian noise to recover a high-quality image. Consequently, the restored image $\hat{\mathcal{X}}$ should be perceptually similar to the original high-quality image $\mathcal{X}$.
 
We define the training pair $\{\mathcal{X},\mathcal{Y}\}$, where $\mathcal{X}$ represents the ground truth (HR) image, and $\mathcal{Y}$ denotes the degraded LR image. The overall training process is shown in Fig.~\ref{fig:main_diag}. Our approach utilizes a pretrained text-to-image-based Stable Diffusion model in conjunction with ControlNet as the backbone architecture for training. Following the methodology introduced in ControlNet, we replicate the downsampling blocks of the frozen Stable Diffusion model and employ them as trainable modules. These trainable modules are connected to their corresponding frozen counterparts using zero-convolution blocks, ensuring smooth information flow while enabling targeted updates. The diffusion process occurs in the latent space, where the Stable Diffusion encoder module transforms the HR image into its latent representation as $\mathrm{Z}_0 = \mathcal{E}_{\text{VAE}}(\mathcal{X})$. The output of the diffusion process is mapped back to the pixel space through the decoder module. During training, only the attention layers of the down, mid, and up blocks of the pretrained Stable Diffusion UNet are updated, allowing the model to effectively learn restoration-specific features while preserving the general structure of the pretrained network. For handling the encoder hidden states of both the ControlNet and UNet, we employ a DINO~\cite{caron2021emerging} module capable of extracting robust image representations. To disentangle the model from text-based features, we incorporate a DINO module to condition the diffusion process using meaningful high-quality features, defined as $c_d =\mathcal{E}_{\text{DINO}}(\mathcal{Y})$. Additionally, the conditioning image for the ControlNet is processed through a lightweight network $\mathcal{E}$, which extracts meaningful RGB feature embeddings $c_f = \mathcal{E}(\mathcal{Y})$ from the low-quality input image. This embedding is then added to the ControlNet's noisy latent input via a zero-convolution layer, further enhancing the conditioning mechanism.

The latent diffusion MSE loss is obtained as
\begin{equation}
    \mathcal{L}_{\epsilon} = \mathbb{E}_{c_d, t, c_f,\epsilon\sim \mathcal{N}(0, I)}[|| \epsilon -\epsilon_{\theta}( z_t, t, c_d, c_f)
    ||_2^2]
    \label{eq:loss_denoising}
\end{equation}



\begin{figure*}
  \centering
   \includegraphics[width=1.0\linewidth]{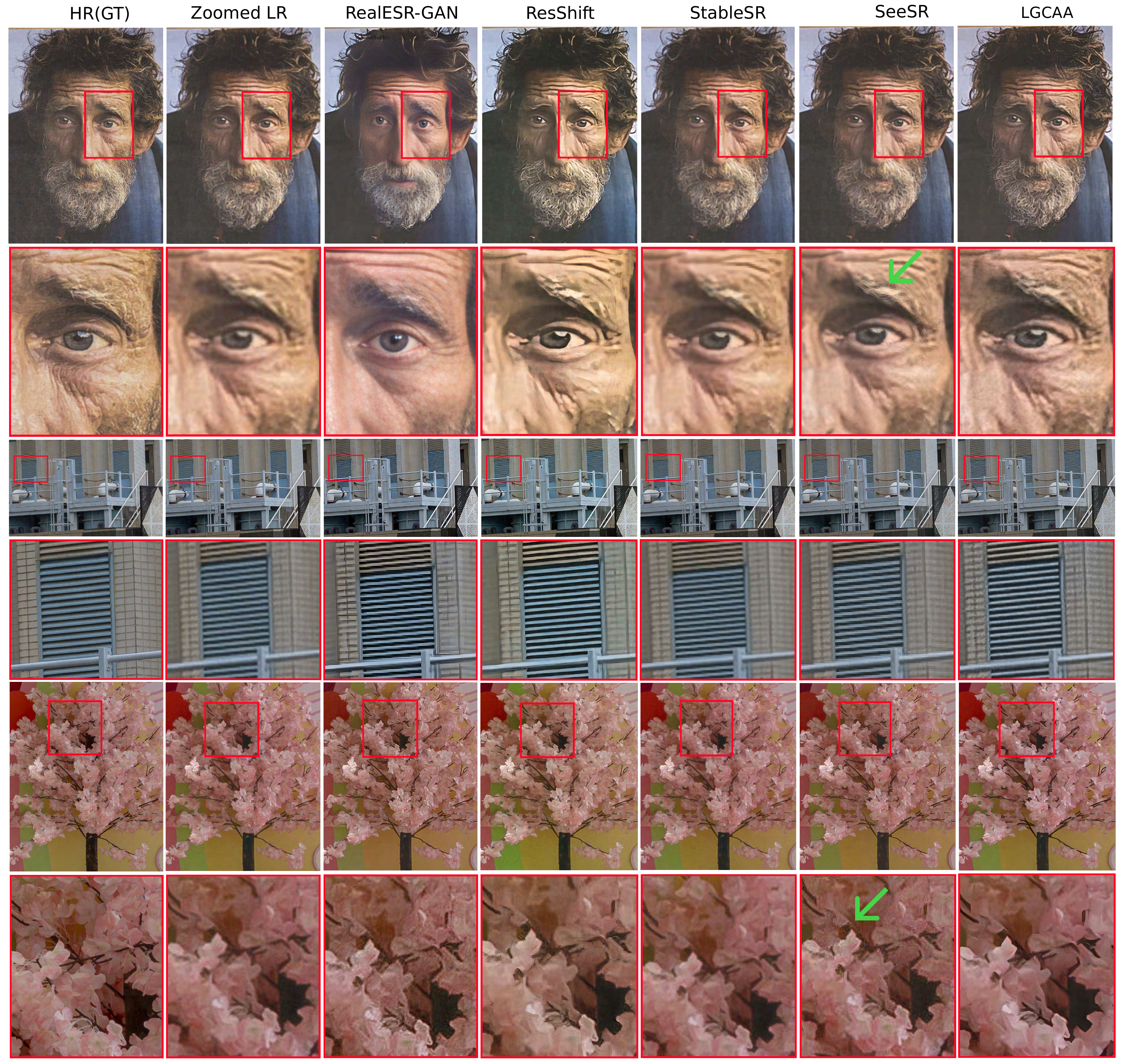}

   \caption{Comparison of our method with other methods on RealSR dataset. It can be seen that our method looks more close to the original ground truth without enhancing any unintended artifacts. Please zoom in for better view.}
   \label{fig:sr_comparison}
\end{figure*}

\subsection{Local and Global Context Aware Attention (LGCAA)}
Given the effectiveness of attention mechanisms in preserving object properties and enhancing image reconstruction, we introduce a Local and Global Context-Aware Self-Attention mechanism. This approach computes self-attention over both local and global image regions to capture fine-grained details and long-range dependencies. 

To effectively model local attention, we first project the input features into a global embedding space and compute the corresponding attention scores. Specifically, given an input feature map $\mathcal{S}$, we first reshape it to $\mathcal{S}'$ and normalize it to obtain $\text{LN}(\mathcal{S}')$. Subsequently, we compute the query  matrices. For local attention, we normalize the $\mathrm{Q}$, $\mathrm{K}$ and $\mathrm{V}$ matrices and compute the local attention weights as, $\mathcal{A}_\text{L} = \texttt{softmax} (\frac{\mathrm{Q}\mathrm{K}^T}{\sqrt{d_k}})$, where $d_k$ is the dimension of each attention heads. The local attention output is then obtained as: $\mathcal{A}_L\mathrm{V}$. Here $\mathcal{A}_\text{L}$  captures the local interactions between neighboring pixels. Finally, the local attention output undergoes normalization to ensure stable feature representation.

To incorporate global attention, we project the locally attended feature map $\hat{\mathcal{S}}_\text{L}$ into a global embedding space and normalize it. The global attention is then computed on this projected representation to capture long-range dependencies within the image. Subsequently after computing the Global attention $\mathcal{S}_\text{G} = \text{GA}\{\hat{\mathcal{S}}_\text{L}\}$ it is normalized and finally reshaped back to the original dimension. This attention mechanism helps to keep a well balance between the high frequency growing components from local to the global region. While the local attention will enhance local interesting region by sharpening consequently if found highly intended for the global value the clamping and the normalization in the global attention suppresses that abnormal growth of high frequency part. We now present the overall pipeline in its mathematical formulation.

\subsubsection{Mathematical Formulation}
\begin{itemize}
    \item \textbf{Input Transformation:}  
    Let the input feature map be denoted as $\mathcal{S} \in \mathbb{R}^{\text{B} \times \text{C} \times \text{H} \times \text{W}}$.  
    The input is first reshaped and normalized as  
    $\mathcal{S}' = \texttt{reshape}(\mathcal{S}) \in \mathbb{R}^{\text{B} \times (\text{HW}) \times \text{C}}$,  
    followed by $\hat{\mathcal{S}} = \text{LN}_1(\mathcal{S}')$.

    \item \textbf{$\mathrm{Q}$, $\mathrm{K}$, and $\mathrm{V}$ Computation:}  
    The query, key, and value matrices are computed as  
    $\mathrm{Q}, \mathrm{K}, \mathrm{V} = \text{Proj}_{\text{in}}(\hat{\mathcal{S}})$,  
    where $\mathrm{Q}, \mathrm{K}, \mathrm{V} \in \mathbb{R}^{\text{B} \times (\text{HW}) \times \text{C}}$.  
    These tensors are reshaped for multi-head attention, yielding  
    $\mathrm{Q}, \mathrm{K}, \mathrm{V} \in \mathbb{R}^{\text{B} \times \text{HW} \times \text{num heads} \times (\text{C}/\text{num heads})}$,  
    and subsequently rearranged to  
    $\mathrm{Q}, \mathrm{K}, \mathrm{V} \in \mathbb{R}^{\text{B} \times \text{num heads} \times \text{HW} \times (\text{C}/\text{num heads})}$.

    \item \textbf{Local Attention:}  
    The queries and keys are normalized as  
    $\mathrm{Q} = \frac{\mathrm{Q}}{\max(|\mathrm{Q}|, \epsilon)}$,  
    $\mathrm{K} = \frac{\mathrm{K}}{\max(|\mathrm{K}|, \epsilon)}$.  
    The local attention scores are then computed as  
    $\hat{\mathcal{A}}_\text{L} = \texttt{softmax}\left(\frac{\mathrm{QK}^\text{T}}{\sqrt{d_k}}\right)$,  
    where $d_k = \text{C}/\text{num heads}$ denotes the dimension per attention head.  
    The local attention output is given by  
    $\mathcal{S}_\text{L} = \mathcal{A}_\text{L} \mathrm{V}$.  
    This output is reshaped as $\mathcal{S}'_\text{L} = \texttt{reshape}(\texttt{permute}(\mathcal{S}_\text{L}))$,  
    projected via $\mathcal{S}''_\text{L} = \text{Proj}_{\text{out}}(\mathcal{S}'_\text{L})$,  
    and normalized as $\hat{\mathcal{S}}_\text{L} = \text{LN}_2(\mathcal{S}''_\text{L})$.

    \item \textbf{Global Attention:}  
    Global attention is applied as  
    $\mathcal{S}_\text{G} = \text{GA}\{\hat{\mathcal{S}}_\text{L}\}$,  
    followed by value clamping  
    $\mathcal{S}_\text{G} = \text{clamp}(\mathcal{S}_\text{G}, -1, 1)$  
    to mitigate potential NaN values.
    \item \textbf{Reshaping to Original Dimensions:}  
    The global attention output is rearranged as  
    $\mathcal{S}'_\text{G} = \texttt{permute}(\texttt{reshape}(\mathcal{S}_\text{G}))$,  
    where $\mathcal{S}'_\text{G} \in \mathbb{R}^{\text{B} \times \text{C} \times \text{H} \times \text{W}}$.  
    Layer normalization is applied as  
    $\mathcal{S}''_\text{G} = \text{LN}_3(\mathcal{S}'_\text{G})$,  
    followed by an MLP and flattening to produce the final output  
    $\mathcal{Y} = \text{MLP}(\texttt{flatten}(\mathcal{S}''_\text{G}))$,  
    where $\mathcal{Y} \in \mathbb{R}^{\text{B} \times \text{C} \times \text{H} \times \text{W}}$.

    \item \textbf{Final Output:} The complete formulation of the output is thus given by 
    \begin{align*}    
    \mathcal{Y} =  
    \text{MLP}(\text{LN}_3(\text{GA}(\text{LN}_2(\text{Proj}_{\text{out}}( 
    \texttt{softmax}(
    \frac{\mathrm{QK}^\text{T}}{\sqrt{d_k}})\mathrm{V})))
    \end{align*}
\end{itemize}

\subsection{Distribution and Perceptual Aligned Conditioning Module (DPACM)}
To enhance and preserve the pixel-level details of the LR image while guiding its transformation toward HR reconstruction, we incorporate Wasserstein Distance  Loss and Perceptual Loss in our training framework. The Wasserstein Distance Loss aims to align the overall latent space pixel distribution of the generated HR image with that of the ground-truth image, ensuring a more realistic reconstruction. This helps to reduce the color shifts and also preserve structural similarity. And hence the LR pixel distribution remains much closer to the HR pixel distribution. Meanwhile, the Perceptual Loss enforces perceptual similarity between the generated HR image and the ground truth, preserving fine details and structural consistency. Furthermore, the ControlNet conditioning vector $c_f$ contains rich pixel-level details extracted from the conditioning input, making it a valuable feature representation for guiding the diffusion process. To effectively leverage this, we refine the diffusion process by incorporating Wasserstein Loss and perceptual Loss with respect to these embedding vectors. To ensure compatibility with the RGB space, we use a convolutional layer that transforms the conditioning embedding $c_f$ into an RGB representation of the LR image: $\mathcal{X}_{RGB} = \mathrm{conv2D}(c_f)$.

\begin{equation}
    \mathcal{L}_{\text{perceptual}}(\mathcal{X}, \mathcal{X}_{\mathrm{RGB}}) = ||\phi_l(\mathcal{X}_{\mathrm{RGB}})-\phi_l(\mathcal{X})||_2^2
    \label{eq:loss_percep}
\end{equation}
Where we take $\phi(\cdot)$ as an alexnet~\cite{krizhevsky2012imagenet} as the feature extractor. In addition we use Wasserstein distance as over the infinimum of the joint pixel distribution $\mathcal{X}$ and $\mathcal{X}_{\mathrm{RGB}}$ as,
\begin{align*}
    \mathcal{L}_{\text{distribution}}(\mathcal{X}, \mathcal{X}_{\mathrm{RGB}}) = \underset{\gamma \in\Pi(\mathcal{X}, \mathcal{X}_{\mathrm{RGB}})}{\mathrm{inf}} \mathbb{E}_{(i,j)\sim \gamma}[d(i,j)]\\ 
\end{align*}

Here $\Pi(\mathcal{X}, \mathcal{X}_{\mathrm{RGB}})$ is the joint distribution of $\mathcal{X}$ and $\mathcal{X}_{\mathrm{RGB}}$ whose marginals will give the individual distributions and $d(i,j)$ represents the ground cost of transporting mass from pixel $i$ of $\mathcal{X}$ to pixel $j$ of $\mathcal{X}_{\mathrm{RGB}}$. We employ the Earth Mover’s Distance (Wasserstein-1 distance), to quantify the discrepancy, where the ground cost is defined as the absolute pixel-wise difference, given by:
\begin{align*}
    d(i,j) = |\mathcal{X}_i-\mathcal{X}_{\text{RGB},j}|
\end{align*}
This results in a simplified closed-form expression that is computationally efficient.
\begin{equation}
    \mathcal{L}_{\text{distribution}}(\mathcal{X}, \mathcal{X}_{\mathrm{RGB}}) = \frac{1}{N}\sum_{k=1}^{N}|\mathcal{X}_k-\mathcal{X}_{\text{RGB},k}|
    \label{eq:loss_distri}
\end{equation}
This formulation preserves the geometric interpretation of the Wasserstein distance, or optimal transport—commonly referred to as the Earth Mover’s Distance—while rendering it tractable for pixel-level refinement.

Hence, by combining Equations~\ref{eq:loss_denoising}, \ref{eq:loss_percep}, and \ref{eq:loss_distri}, the overall loss function is expressed as:
\begin{equation}
\mathcal{L}_{\mathrm{LGCAA}} = \mathcal{L}_{\epsilon} + \lambda_l\mathcal{L}_{\text{perceptual}} + \lambda_{w}\mathcal{L}_{\text{distribution}}
\end{equation}
Here $\lambda_l$ and $\lambda_w$ are balance between distribution and perceptual loss terms. 

\begin{figure*}[!ht]
  \centering
   \includegraphics[width=1.0\linewidth]{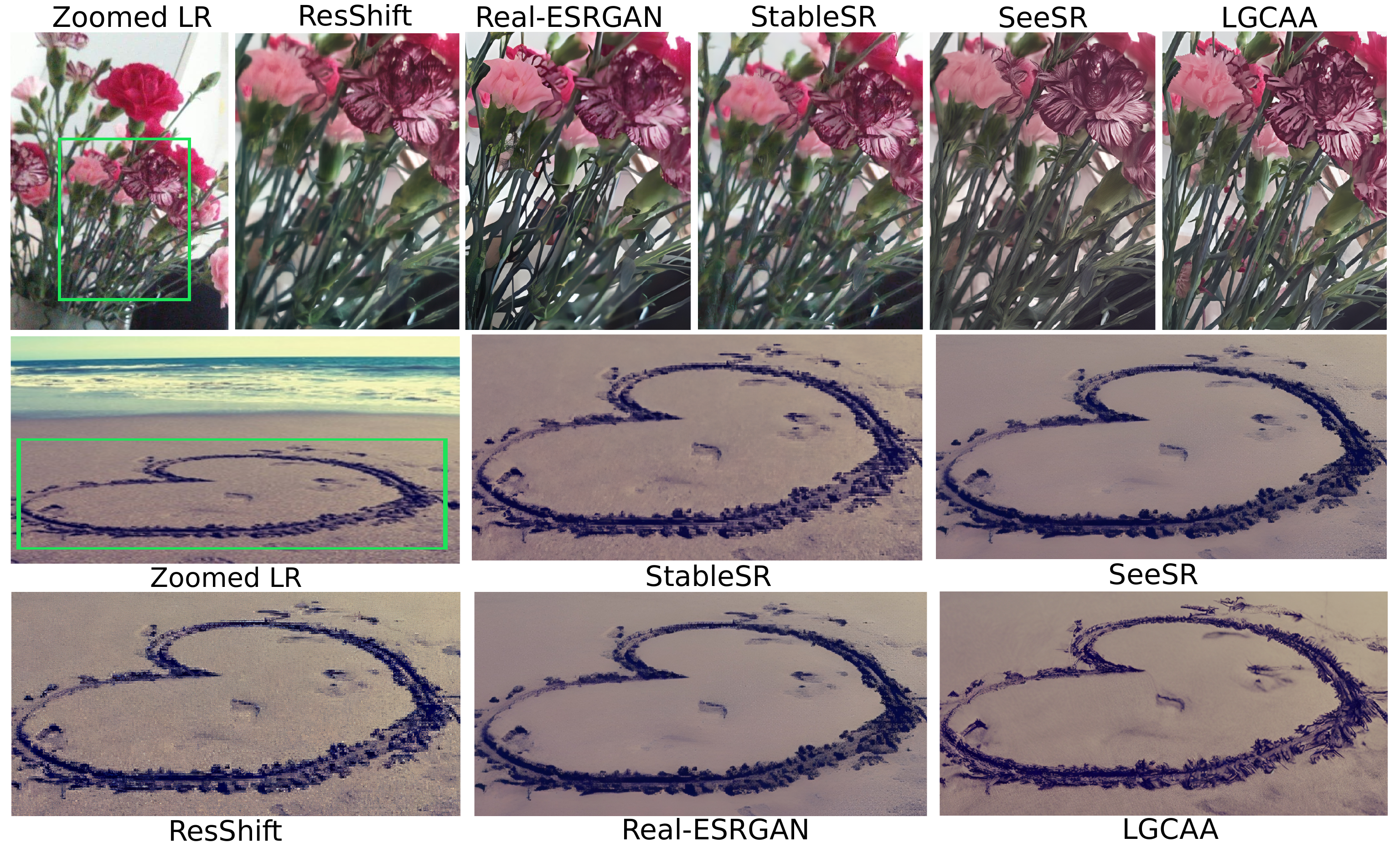}

   \caption{Visual comparison of various methods on a real-world dataset example without any HR reference image. Please zoom in for a better view.}
   \label{fig:supp_1}
\end{figure*}

\begin{figure*}[!htbp]
  \centering
   \includegraphics[width=1.0\linewidth]{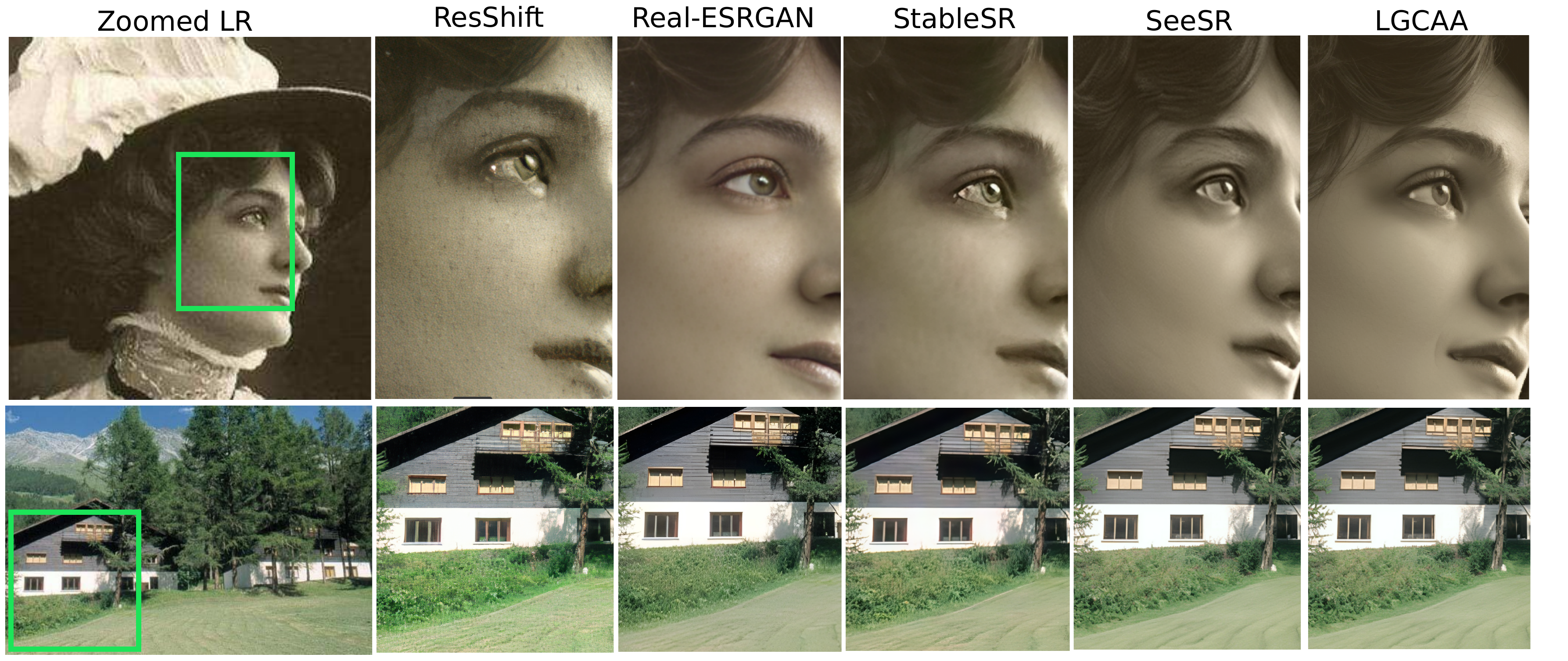}

   \caption{Visual comparison of various methods on a real-world dataset example without any HR reference image. Please zoom in for a better view.}
   \label{fig:supp_2}
\end{figure*}

\section{Experiments}
\label{sec:exp}
To show the effectiveness of LGCAA we show qualitative comparison results and also extensive quantitative results. We show our experiments on the $\times4$ Real-ISR task on the RealSR dataset similar to the existing methods~\cite{wang2021real,zhang2021designing}. 
\begin{figure*}
  \centering
   \includegraphics[width=1.0\linewidth]{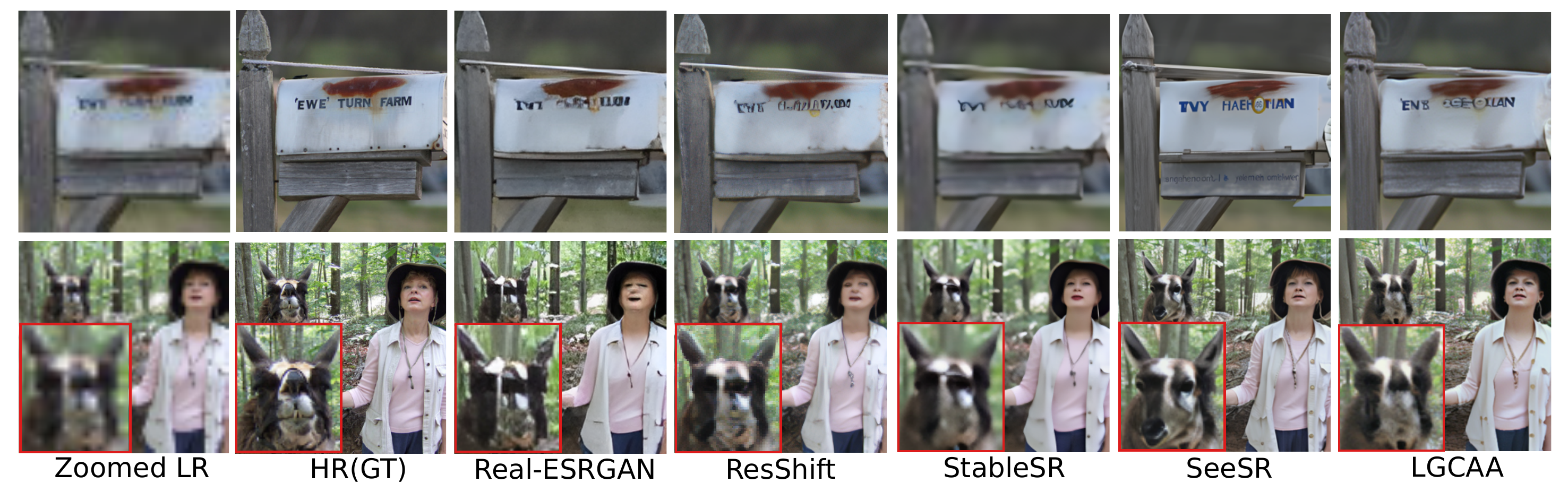}

   \caption{ Comparison of our method with other methods on Imagenet-Test dataset. In the first example, the word `EWE' is not accurately reconstructed by prior methods, whereas LGCAA recovers it with higher fidelity, closely matching the ground truth. In contrast, SeeSR incorrectly reconstructs the text as `TVY', which misrepresents the original content. In the second example, SeeSR alters the face of the pet into that of a deer, deviating significantly from the HR reference despite improving image sharpness. Although LGCAA’s output may appear slightly blurry—particularly in regions where the LR input contains minimal detail—it preserves the original content without introducing semantic deviations, a challenge also observed in other methods. Please zoom in for better view.}
   \label{fig:sr_comparison_imgnt}
\end{figure*}

\begin{table*}
  \centering
  \scalebox{0.9}{
  \begin{tabular}{m{1.25cm} | c |m{1.2cm} m{1.29cm} m{1cm} m{1.1cm} | m{1cm} m{1.4cm} m{1.28cm} m{1.28cm} m{1.2cm} m{1.2cm}} 
    \hline
    \textbf{Datasets} & \textbf{Metrics} & \textbf{BSRGAN} \cite{zhang2021designing} &\cite{wang2021real}\textbf{Real-ESRGAN} & \textbf{LDL} \cite{liang2022details} & \textbf{DASR} \cite{liang2022efficient} & \textbf{LDM} \cite{rombach2022high}& \textbf{StableSR} \cite{wang2024exploiting} &\textbf{ResShift} \cite{yue2023resshift} & \textbf{PASD} \cite{yang2024pixel}& \textbf{SeeSR} \cite{wu2024seesr} & \textbf{LGCAA}\\
    \hline 
     & PSNR $\uparrow$& 21.87 & \underline{21.94} & 21.52 & 21.72 & 21.26 & 20.84 & 21.75 & 20.77 & 21.19 & \textbf{21.98}\\
     & SSIM $\uparrow$& 0.5539 & \underline{0.5736} & 0.5690 & 0.5536 & 0.5239 & 0.4887 & 0.5422 & 0.4958 & 0.5386 & \textbf{0.5745}\\
    & LPIPS $\downarrow$& 0.4136 & 0.3868 & 0.3995 & 0.4266 &0.4154 & 0.4055 & 0.4284 & 0.4410 & \underline{0.3843} & \textbf{0.3821}\\
    & MUSIQ $\uparrow$& 59.11 & 58.64 & 57.90 & 54.22 & 56.32 & 62.95 & 58.23 & 65.23 & \textbf{68.33} & \underline{66.17}\\
    \textbf{DIV2K-}& CLIPIQA $\uparrow$& 0.5183 & 0.5424 & 0.5313 & 0.5241 & 0.5695 & 0.6486 & 0.5948 & \underline{0.6799} & \textbf{0.6946} & 0.6715\\
    \textbf{Val}& DISTS $\downarrow$& 0.2737 & 0.2601 & 0.2688 & 0.2688 & 0.2500 & 0.2542 & 0.2606 & 0.2538 & \underline{0.2257} & \textbf{0.2215}\\
    & FID $\downarrow$ & 64.28 & 53.46 & 58.94 & 67.22 & 41.93 & \underline{36.57} & 55.77 & 40.77 & \textbf{31.93} & 38.72\\
    & NIQE $\downarrow$& 4.7615 & 4.9209 & 5.0249 & 4.8596 & 6.4667 & \textbf{4.6551} & 6.9731 & 4.8328 & 4.9275 & \underline{4.7157}\\
    & MANIQA $\uparrow$& 0.4834 & 0.5251 & 0.5127 & 0.4346 & 0.5237 & 0.5914 & 0.5232 & 0.6049 & \textbf{0.6198} & \underline{0.6175}\\
    \hline
    & PSNR $\uparrow$& 26.39  & 25.69 & 25.28 & \underline{27.02} & 25.48 & 24.70 & 26.31 & 25.18 & 24.29 & \textbf{27.05}\\
    & SSIM $\uparrow$& 0.7654 & 0.7616 & 0.7567 & \underline{0.7708} & 0.7148 & 0.7085 & 0.7421 & 0.6630 & 0.7216 & \textbf{0.7715}\\
    & LPIPS $\downarrow$& \textbf{0.2670} & 0.2727 & 0.2766 & 0.3151 & 0.3180 & 0.3018 & 0.3460 & 0.3435 & 0.3009 & \underline{0.2715} \\
    & MUSIQ $\uparrow$& 63.21 & 60.18 & 60.82 & 40.79 & 58.81 & 65.78 & 58.43 & 68.69 & \textbf{69.77} & \underline{68.95}\\
    \textbf{RealSR}& CLIPIQA $\uparrow$& 0.5001 & 0.4449 & 0.4477 & 0.3121 & 0.5709 & 0.6178 & 0.5444 & 0.6590 & \textbf{0.6612} &  \underline{0.6595}\\
    & DISTS $\downarrow$& 0.2121 & \textbf{0.2063} & 0.2121 & 0.2207 & 0.2213 & 0.2135 & 0.2498 &0.2259&0.2223&\underline{0.2097}\\
    & FID $\downarrow$ & 141.28& 135.18 & 142.71 & 132.63 & 132.72 & 128.51 & 141.71 & 129.76 & \textbf{125.55} & \underline{128.34} \\
    & NIQE $\downarrow$& 5.6567 & 5.8295 & 6.0024 & 6.5311& 6.5200 & 5.9122 & 7.2635 & \textbf{5.3628} & \underline{5.4021} & 5.5176\\
    & MANIQA $\uparrow$& 0.5399 & 0.5487 & 0.5485 & 0.3878 & 0.5423 & 0.6221 & 0.5285 & \textbf{0.6493} & \underline{0.6442} & 0.6433\\
    \hline
    & PSNR $\uparrow$& 28.75 & 28.64 &  28.21 & \underline{29.77} & 27.98 & 28.13 & 28.46 & 27.00 & 28.17 & \textbf{29.82} \\
    & SSIM $\uparrow$& 0.8031 & 0.8053 &  0.8126 & \underline{0.8264} & 0.7453 & 0.7542 & 0.7673 & 0.7084 & 0.7691 & \textbf{0.8271}\\
    & LPIPS $\downarrow$ & 0.2883 & 0.2847 & \underline{0.2815} & 0.3126 & 0.3405 & 0.3315 & 0.4006 & 0.3931 &0.3189 & \textbf{0.2809}\\
    & MUSIQ $\uparrow$& 57.14 & 54.18 & 53.85 & 42.23 & 53.73 & 58.42 & 50.60 & \underline{64.81} & \textbf{64.93} & 64.68 \\
    \textbf{DRealSR}& CLIPIQA $\uparrow$& 0.4915 & 0.4422 & 0.4310 & 0.3684 & 0.5706 & 0.6206 & 0.5342 & \underline{0.6773} &\textbf{0.6804} & 0.6695\\
    & DISTS $\downarrow$& 0.2142 & \textbf{0.2089} & 0.2132 & 0.2271 & 0.2259 & 0.2263 & 0.2656 & 0.2515 & 0.2315 & \underline{0.2106}\\
    & FID $\downarrow$ & 155.63 & \underline{147.62} & 155.53 & 155.58 & 156.01 & 148.98 & 172.26 & 159.24 & \textbf{147.39} & 148.54\\
    & NIQE $\downarrow$& 6.5192 & 6.6928 & 7.1298 & 7.6039 & 7.1677 & 6.5354 & 8.1249 & \textbf{5.8595} & 6.3967 & \underline{5.8923}\\
    & MANIQA $\uparrow$& 0.4878 & 0.4907 & 0.4914 & 0.3879 & 0.5043 & 0.5591 & 0.4586 & 0.5850 & \textbf{0.6042} & \underline{0.5932}\\
    \hline
  \end{tabular}}
  \caption{We conduct a quantitative comparison of our approach with state-of-the-art Real-ISR models based on GAN and diffusion frameworks across various datasets. The best-performing method is highlighted in bold, while the second-best result is indicated with an underline.}
  \label{tab:table_our_results}
\end{table*}

\begin{figure}
    \centering
    \includegraphics[width=1.0\linewidth]{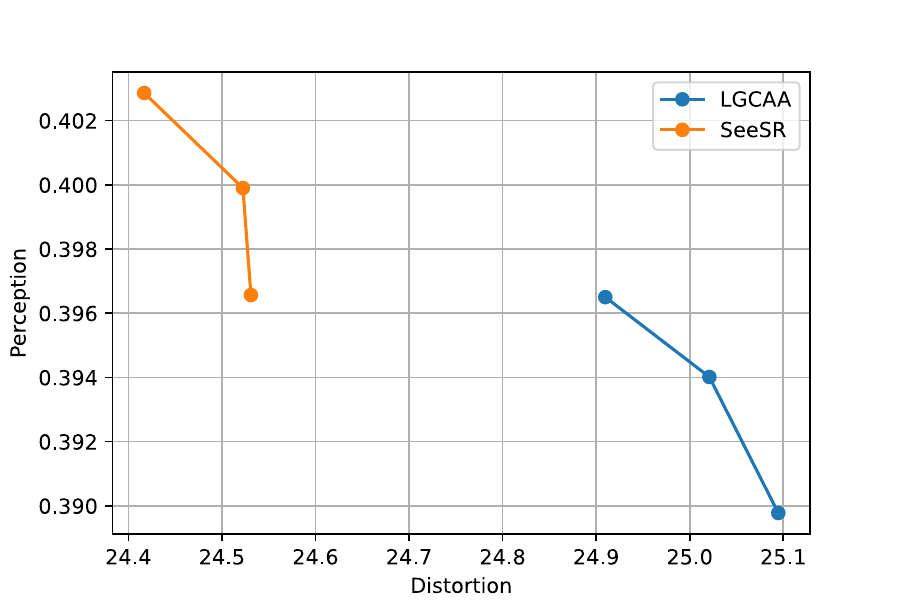}
    \caption{
    The perception-distortion tradeoff of LGCAA is compared with SeeSR~\cite{wu2024seesr}, where perception and distortion are measured using LPIPS and PSNR, respectively}
    \label{fig:percep-distor}
\end{figure}

\begin{figure*}[!ht]
  \centering
   \includegraphics[width=1.0\linewidth]{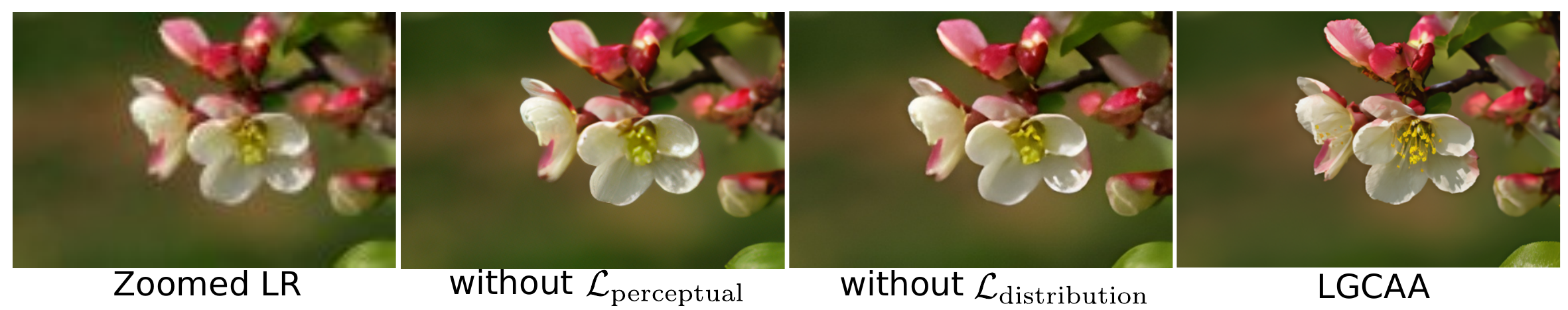}

   \caption{Significance of Various Loss Terms in DPACM module.}
   \label{fig:supp_3}
\end{figure*}

\begin{figure*}[!ht]
  \centering
   \includegraphics[width=1.0\linewidth]{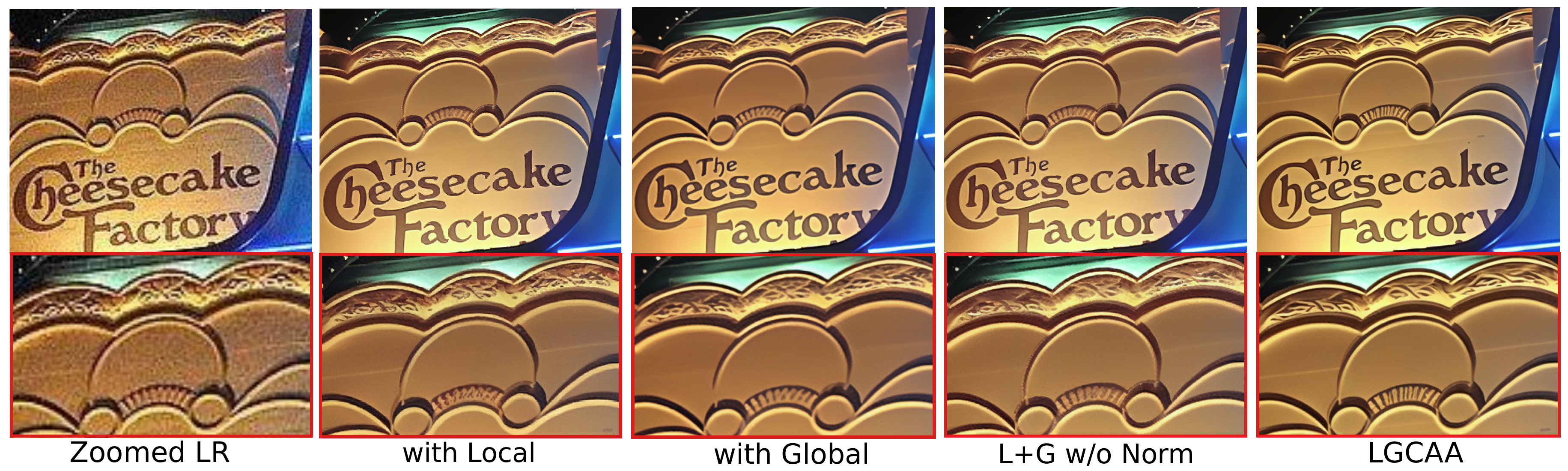}

   \caption{Significance of Various Loss Terms in LGCAA module.}
   \label{fig:supp_4}
\end{figure*}

\begin{figure*}
  \centering
   \includegraphics[width=1.0\linewidth]{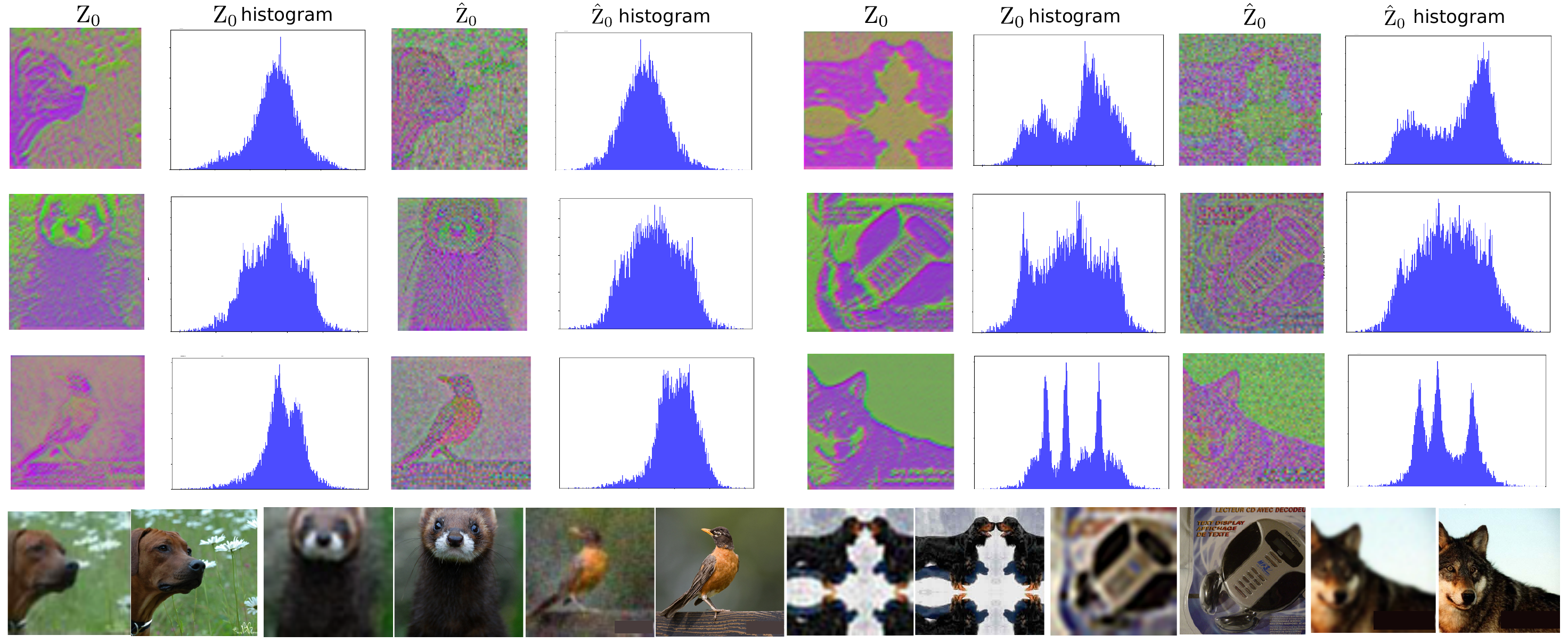}

   \caption{We present a histogram comparison between the predicted latent representation, $\hat{\mathrm{Z}}_0$, and the original latent representation, $\mathrm{Z}_0$. In the histogram plots, the horizontal axis represents the normalized pixel intensity, while the vertical axis denotes the frequency of occurrence. The first three rows illustrate the histogram comparisons between $\hat{\mathrm{Z}}_0$ and $\mathrm{Z}_0$, whereas the last row displays the corresponding LR-HR images.}
   \label{fig:hist_comparison}
\end{figure*}

\subsection{Experimental settings}
\textbf{Training details:}
HR images with a resolution of $256\times256$ in our training dataset are randomly cropped from the ImageNet training set~\cite{deng2009imagenet}, following the approach of LDM~\cite{rombach2022high}. To generate LR images, we adopt the degradation pipeline of RealESRGAN~\cite{wang2021real}. Our Real-ISR model is based on the pretrained Stable Diffusion 2 (SD-base 2) text-to-image model. We freeze the existing modules of the UNet and optimize the components outlined in Sec. \ref{sec:method}, incorporating ControlNet for additional conditioning. For training, we utilize the Adam~\cite{diederik2014adam} optimizer with $\beta_1 =0.9$ and $\beta_2 =0.999$. The controlled text-to-image model is trained with a batch size of 192 for a total of 150K iterations. The training process is conducted using a single NVIDIA DGX A100 GPU, consuming approximately 24GB of the available 80GB memory. During inference, we employ 40 DDPM sampling steps.

\textbf{Test Dataset:}
We conduct our experiments using four datasets. We utilize the DIV2K dataset~\cite{agustsson2017ntire}, which contains 3K LR-HR image pairs. LR images of size $128\times128$ are generated from the HR images of size $512\times512$ using the same degradation pipeline as in the training dataset. Following~\cite{yue2023resshift}, we utilize the ImageNet-test dataset, which incorporates additional degradation kernels to facilitate evaluation under more severe degradation scenarios. To assess real-world degradations, we incorporate the RealSR dataset~\cite{cai2019toward} and DRealSR~\cite{wei2020component} datasets.

\textbf{Compared methods:} 
We compare our method with GAN-based approaches, including BSRGAN~\cite{zhang2021designing}, RealESRGAN~\cite{wang2021real}, LDL~\cite{liang2022details}, and DASR~\cite{liang2022efficient}, as well as diffusion model-based methods such as LDM~\cite{rombach2022high}, StableSR~\cite{wang2024exploiting}, ResShift~\cite{yue2023resshift}, and SeeSR~\cite{wu2024seesr}. During inference, different methods utilize varying numbers of sampling steps (e.g., LDM uses 1000 steps, while ResShift employs 15 steps). To ensure a fair comparison, we adopt the best-performing step configurations as reported in their respective works.

\textbf{Metrics:}
We utilize five evaluation metrics to compare our method with other state-of-the-art approaches, incorporating both reference-based and no-reference-based metrics. The reference-based metrics include PSNR, SSIM~\cite{wang2004image}, LPIPS~\cite{zhang2018unreasonable}, and DISTS~\cite{ding2020image}. Additionally, we employ no-reference image quality assessment metrics, namely FID~\cite{heusel2017gans}, MUSIQ~\cite{ke2021musiq}, CLIPIQA~\cite{wang2023exploring}, NIQE~\cite{zhang2015feature}, and MANIQA~\cite{yang2022maniqa}, to further evaluate perceptual quality.

\textbf{Quantitative Comparison:}
We present a quantitative comparison of our method against five GAN-based and diffusion-based Real-ISR approaches across three different datasets in Table~\ref{tab:table_our_results}. The results demonstrate that our method consistently outperforms all competing methods in PSNR and SSIM, even surpassing GAN-based approaches. Additionally, it achieves the second-best performance in CLIPIQA and DISTS scores for the RealSR dataset, with a marginal difference from the best-performing method. SeeSR, which leverages semantic and text-based features, excels in CLIPIQA, as well as in FID and MUSIQ scores due to its text-based refinements. Diffusion model-based methods generally perform well in MANIQA and MUSIQ, with our method achieving the second-best results for both metrics on the RealSR dataset. Furthermore, for the RealSR dataset, our approach surpasses the previously second-best GAN-based method by achieving a 1.13\% lower DISTS score, despite GAN-based methods dominating both the best and second-best rankings for this metric. Additionally, our method reduces LPIPS by 0.44\% compared to the second-best GAN-based LPIPS score, where GANs also had the best overall performance. In terms of the DISTS metric, our method performs consistently well across all three datasets, securing either the best or second-best score. These results highlight that our approach achieves highly competitive performance compared to both GAN-based and diffusion model-based methods.

\textbf{Qualitative Comparison:}
We present the results on real-world degradation datasets in Fig.~\ref{fig:sr_comparison}. It is evident that LGCAA produces results that are more realistic and closely aligned with the ground truth image, without introducing unnecessary high-frequency details to artificially enhance image quality. Existing methods often introduce unwanted artifacts into the reconstructed images. GAN-based approaches perform well in generating smooth color transitions for objects; however, upon closer inspection, they tend to introduce excessive smoothing and overly bright colors, leading to unrealistic visual artifacts. Among pretrained diffusion-based models, SeeSR demonstrates strong performance in image restoration and Real-ISR tasks. However, its reliance on text-based prompts can introduce unintended artifacts in the generated images.  For instance, in an image of a plant, the model erroneously introduces nut-shaped objects in black gap regions. This behavior suggests that the text embedding may have misinterpreted the gap as an object rather than empty space. Similarly, in an image of a person, the model adds unwanted extra hairs to the eyebrows, likely due to ambiguities in the text-based conditioning. These findings highlight a limitation of text-guided diffusion models in super-resolution tasks, where the reliance on textual embeddings can lead to hallucinated details that deviate from the original structure of the image. For Real-ESRGAN, StableSR, and ResShift, the overall image quality appears visually promising at first glance. However, when zoomed in, unintended artifacts often become noticeable. In contrast, our method effectively reconstructs super-resolved images while maintaining control over detail generation through local and global context-aware attention. By balancing high and low frequency components, our approach ensures visually appealing results without introducing unwanted artifacts.

Similarly, in the case of highly degraded images from the ImageNet-Test dataset, we present the results in Fig.~\ref{fig:sr_comparison_imgnt}. While other methods struggle to reconstruct images close to the ground truth, LGCAA demonstrates superior performance in preserving the original content. For instance, in the first image, the word `EWE' is not accurately reconstructed by existing methods, whereas LGCAA is able to recover it more effectively. In contrast, SeeSR generates an entirely different word, highlighting the challenge of text-based conditioning in extreme degradation scenarios. In the second example, LGCAA successfully restores finer details in both the human face and the pet’s face, maintaining structural accuracy. In comparison, the GAN-based Real-ESRGAN produces an overly smoothed output that fails to resemble the original facial features. These results demonstrate LGCAA’s ability to recover high-quality details even under severe degradation.
We provide more visual comparison different real world datasets without any HR reference image in Fig.~\ref{fig:supp_1} and Fig.~\ref{fig:supp_2}.

\begin{table}
  \centering
  \scalebox{0.9}{
  \begin{tabular}{c | c| c c c c } 
    \hline
    \textbf{Module} & \textbf{Variant} & \textbf{PSNR} $\uparrow$ & \textbf{LPIPS} $\downarrow$& \textbf{CLIPIQA}$\uparrow$ &\textbf{MUSIQ} $\uparrow$\\
    \hline 
     & No LGCAA& 25.25 &  0.2902 & 0.6321 & 66.73 \\
     & Local& 25.65 & 0.2869 & 0.6388 & 67.17 \\
    LGCAA& Global & 25.52 & 0.2865 & 0.6381 & 67.35\\
    & L+G w/o norm& 26.45 & 0.2872 & 0.6402 & 67.21 \\
    & LGCAA & 27.05 & 0.2715 & 0.6595 & 68.95\\
    \hline
    & No DPACM& 25.45 & 0.2842 & 0.6384 & 66.52\\
    DPACM& Perceptual & 25.78 & 0.2765 & 0.6472 & 67.42 \\
    & Wasserstein& 25.82 & 0.2758 & 0.6457 & 67.54\\
    & DPACM & 27.05 & 0.2715 & 0.6595 & 68.95 \\
    \hline
  \end{tabular}
  }
  \caption{We provide an ablation experiment with different components of the LGCAA and the DPACM module on RealSR dataset.}
  \label{tab:ablation_dpacm_lgcaa}
\end{table}

\begin{table*}[!h]
  \centering
  \scalebox{1.0}{
  \begin{tabular}{c c |c c c c c c c c c} 
    \hline
    $\lambda_l$ & $\lambda_w$ & PSNR $\uparrow$ & SSIM $\uparrow$ & LPIPS $\downarrow$& MUSIQ $\uparrow$ & CLIPIQA $\uparrow$ & DISTS $\downarrow$& MANIQA $\uparrow$ & FID $\downarrow$& NIQE $\downarrow$\\
    \hline 
     2.0& 0.1& 26.27 & 0.7478 & 0.2852 & 67.22 & 65.26 & 0.2372 & 0.6364 & 129.19 & 5.252\\
     2.0& 0.2& 26.59 & 0.7516 & 0.2880 & 67.36 & 65.59 & 0.2364 & 0.6315 & 129.56 & 5.176\\
    2.0& 0.3& 26.71 & 0.7528 & 0.2835 & 67.72 &65.92 & 0.2357 & 0.6357 & 128.95 & 5.261\\
    1.0& 0.1& 25.16 & 0.7464 & 0.2957 & 66.27 & 66.12 & 0.2265 & 0.6283 & 130.33 & 5.267\\
    1.0& 0.2& 25.31 & 0.7424 & 0.2913 & 66.41 & 66.35 & 0.2263 & 0.6279 & 130.16 & 5.282\\
    1.0 & 0.3& 25.54 & 0.7421 & 0.2968 & 66.78 & 66.85 & 0.2216 & 0.6248 & 129.87 &5.262\\
    \hline
  \end{tabular}}
  \caption{Quantitative evaluation across various metrics with varying weightings of $\lambda_l$ and $\lambda_w$.}
  \label{tab:table_supp}
\end{table*}

\textbf{Perception distortion trade-off:}
The perception-distortion trade-off in Real-ISR characterizes the balance between fidelity to the ground truth and the perceptual quality of the super-resolved image. This trade-off is critical in image restoration, as enhancing high-frequency details to improve perceptual quality and reduce blurriness may lead to decreased accuracy, and vice versa. In diffusion-based models, increasing the number of sampling steps can improve alignment with the ground truth but may degrade perceptual quality by introducing blurriness. Since SeeSR adopts the same sampling strategy to ensure a fair comparison, we present the perception-distortion curve for both LGCAA and SeeSR in Fig.~\ref{fig:percep-distor}. The evaluation is conducted with DDPM sampling steps of 30, 40, and 50, where perceptual mismatch is quantified using the LPIPS loss, while distortion is measured in terms of PSNR (dB). We see that LGCAA always below and on the right side indicating well balance between perception and distortion and superior behaviour.

\textbf{Ablation on the DPACM and LGCAA Module:}
In the DPACM module, we incorporate both Wasserstein loss and LPIPS loss to enhance super-resolution performance. We evaluate the effect of these losses by adjusting their respective weightings, $\lambda_w$ and $\lambda_l$, and present the corresponding results in Table~\ref{tab:table_supp}. Furthermore, we provide qualitative comparisons of super-resolution outputs under three conditions: one where each loss is individually removed and another where both losses are incorporated, as shown in Figure~\ref{fig:supp_3}. Our observations indicate that employing only Wasserstein loss, as opposed to using only LPIPS loss, introduces finer details but also leads to the generation of certain artifacts. Conversely, relying solely on LPIPS loss results in an overly smooth reconstruction. Therefore, we incorporate both losses to achieve a balance, where the Wasserstein loss enhances visual fidelity, while LPIPS loss mitigates unintended artifacts.

In addition, we present an ablation study on the LGCAA and DPACM modules, as summarized in Table~\ref{tab:ablation_dpacm_lgcaa}.
For LGCAA, we report results under the following settings: standard self-attention training, local attention only, global attention only, combined local and global attention without the normalization layer after merging, and the complete proposed LGCAA module.
We also provide qualitative comparisons of local and global attention outputs in Figure~\ref{fig:supp_4}.
For the DPACM module, we conduct experiments using only the perceptual loss, only the Wasserstein loss, and the combination of both within the DPACM framework.

\textbf{Preserving Histogram consistency:}
Despite alterations in the histogram at the latent level, the structural similarity between the pixel-space representation and the latent-space features ensures a close correspondence between their histograms. Consequently, preserving the distribution of $\mathrm{Z}_0$ is crucial for retaining information. To achieve this, we employ the Wasserstein distance to minimize the discrepancy in pixel distributions between the LR and HR images. Additionally, the Local-Global Context-Aware Attention (LGCAA) module aids in preserving structural integrity and mitigating color shifts in the latent space. As a result, the predicted $\hat{\mathrm{Z}}_0$ distribution closely aligns with the initial $\mathrm{Z}_0$, ensuring consistency. This preservation of the histogram in latent space, as demonstrated in Fig.~\ref{fig:hist_comparison}, further supports the structural fidelity of the reconstructed image.

\section{Conclusions}
In this work, we have introduced an efficient real-world image super-resolution method that effectively enhances the original content while maintaining visually coherent results. Our approach is designed to preserve the integrity of the original image without introducing unnecessary details that may lead to unwanted artifacts. Since high-frequency components contribute to finer details, an excessive emphasis on them can introduce distortions upon closer inspection. To address this, our method carefully balances high- and low-frequency components, ensuring improved visual quality while preventing the generation of unintended artifacts.

\bibliographystyle{IEEEtran}
\bibliography{biblio}

\vfill

\vfill

\end{document}